%% file: paper.tex
\documentclass[sigconf]{acmart}
\usepackage[ruled]{algorithm2e}

\AtBeginDocument{%
  \providecommand\BibTeX{{%
    \normalfont B\kern-0.5em{\scshape i\kern-0.25em b}\kern-0.8em\TeX}}}

\setcopyright{acmcopyright}
\copyrightyear{2022}
\acmYear{2022}

\acmConference[KDD '22 MLG]{28th ACM SIGKDD Conference on Knowledge Discovery and Data Mining, Workshop on Machine Learning in Graphs}{Aug 15,
  2022}{Washington D.C., USA}
\acmDOI{}
\acmPrice{}
\acmISBN{}

\begin{document}

%%
%% The "title" command has an optional parameter,
%% allowing the author to define a "short title" to be used in page headers.
\title{OASYS: Domain-Agnostic Automated System for Constructing Knowledge Base from Unstructured Text}

%%
%% The "author" command and its associated commands are used to define
%% the authors and their affiliations.
%% Of note is the shared affiliation of the first two authors, and the
%% "authornote" and "authornotemark" commands
%% used to denote shared contribution to the research.
\author{Minsang Kim\textsuperscript{\rm*}, Sang-Hyun Je\textsuperscript{\rm*}, Eunjoo Park}
\thanks{* Both authors contributed equally to this research.}
\email{{lucas.ai, dan.j, ellie.s}@kakaoenterprise.com}
\affiliation{%
  \institution{Kakao Enterprise}
  \streetaddress{P.O. Box 1212}
  \city{Pangyo}
  \state{Gyeonggi-do}
  \country{South Korea}
}

%%
%% By default, the full list of authors will be used in the page
%% headers. Often, this list is too long, and will overlap
%% other information printed in the page headers. This command allows
%% the author to define a more concise list
%% of authors' names for this purpose.
\renewcommand{\shortauthors}{Minsang Kim and Sang-Hyun Je et al.}

%%
%% The abstract is a short summary of the work to be presented in the
%% article.
\begin{abstract}
In recent years, creating and managing knowledge bases have become crucial to the retail product and enterprise domains. We present an automatic knowledge base construction system that mines data from documents. This system can generate training data during the training process without human intervention. Therefore, it is domain-agnostic trainable using only the target domain text corpus and a pre-defined knowledge base.
This system is called OASYS and is the first system built with the Korean language in mind. In addition, we also have constructed a new human-annotated benchmark dataset of the Korean Wikipedia corpus paired with a Korean DBpedia to aid system evaluation. The system performance results on human-annotated benchmark test dataset are meaningful and show that the generated knowledge base from OASYS trained on only auto-generated data is useful. We provide both a human-annotated test dataset and an auto-generated dataset.\footnote{https://github.com/kakaoenterprise/OASYS}
\end{abstract}
%%
%% The code below is generated by the tool at http://dl.acm.org/ccs.cfm.
%% Please copy and paste the code instead of the example below.https://www.overleaf.com/download/project/61d7b4b3bf10035f40853ed0/build/18110aa707b-c23287fb7293654d/output/output.pdf?compileGroup=standard&clsiserverid=clsi-pre-emp-e2-c-e-x8w8&popupDownload=true
%%

%%
%% This command processes the author and affiliation and title
%% information and builds the first part of the formatted document.

\maketitle
\input{intro}
\input{method}

\input{dataset}
\input{experiment}

\input{conclusion}

\bibliographystyle{ACM-Reference-Format}
\bibliography{ref}

\appendix
\end{document}

%% file: intro.tex
\section{Introduction}
In the past decade, Bing, Google~\cite{dong2014knowledge}, and Apple~\cite{peng2019improving} have successfully constructed large-scale knowledge bases (KB) for search and question answering. More recently, the knowledge base construction has expanded to a wider range of domains.
In the retail product domain, Amazon~\cite{Dong2020AutoKnowSK} has proposed AutoKnow which is a self-driving system for building a product KB. Meanwhile, Walmart and Alibaba have also built and used product graphs. Rich product knowledge can significantly improve product search, recommendation, and navigation experiences.
In the enterprise domain, Mircosoft~\cite{chai2021automatic} presented an automatic knowledge base construction system without enterprise customization. Since the enterprise domain does not require extreme accuracy like other search engines, it can build fully automatic systems that rely heavily on NLP techniques. This system helps employees better find and explore domain knowledge. As KB are used in more and more domains, there is growing interest in constructing KB with minimal human intervention.\\
\indent In this paper, we present a domain-agnostic knowledge base construction system, which automatically extracts knowledge from documents. This system, which we call OASYS, relies on self-supervised learning and distant supervision to alleviate the burden of manual training data creation while all processes including data generation and model training are performed together during training. Therefore, it only needs the target domain text corpus and pre-defined KB for training and is domain-agnostic trainable.\\
\indent OASYS consists of two main components: 1) the entity linking component and 2) the relation extraction component. The entity linking component, called the entity linker (EL), processes named entity disambiguation. Moreover, the relation extraction component, called the relation extractor (RE), extracts relational information between entity pairs. These respective components are applied sequentially to allow OASYS to extract RDF triples (subject, predicate, object) from raw texts.
For example, given the sentence \textit{``Robert Downey Jr. starred in Avengers: Endgame.''}, EL extracts \textit{``Robert Downey Jr.''} as an entity ID and \textit{``Avengers: Endgame''} as another entity ID in pre-defined KB. Then, RE takes the information and identifies \textit{``starred\_in''} as the relation in between \textit{``Robert Downey Jr.''} and \textit{``Avengers: Endgame''} entities.\\
\indent To be specific, OASYS has essential three features. \textbf{1)} First, OASYS performs the data generation processes that automatically extract labeled data for training without human intervention. EL is a self-supervised training system that performs both data generation and model training processes. For automatic training data generation, EL constructs a sub-graph from all combinations of entity candidates and computes the number of respective connections. Afterward, EL selects the most connected entity to tag the label. Entities extracted in this way have a very low error rate. RE uses distant supervision to generate labels automatically. Though the distant supervision does not require large amounts of manual annotations, it yields highly incomplete and noisy data. To alleviate this problem, RE adopts a multi-instance multi-label learning method. \textbf{2)} As a second feature, OASYS fuses the NLP models and the structured information of the pre-defined KB to improve the system performance. NLP models which extract information from raw texts have limited use since few models can achieve meaningful precision for serving application service. Thus, EL uses results from both the logical sub-graph-based entity linker and the context-based entity linker in the entity disambiguation method. In addition, RE uses Triple Validation that can filter out non-sense triples that do not exist in the pre-defined KB. \textbf{3)} A third feature is that while most of knowledge base construction systems support the English language, OASYS is the first automatic knowledge base construction system for the Korean language. For evaluating system performance in the Korean language, we provide a new human-annotated benchmark dataset of the Korean Wikipedia corpus each paired with a Korean DBpedia~\cite{lehmann2015dbpedia}.
Moreover, we also provide a dataset of over 458K automatically generated data for model training in OASYS. This dataset, with triple and sentence pairs, are essential for a variety of NLP problems, included relation extraction, question answering, and knowledge base-to-text generation.\\
\\
Our contributions are summarized as follows:
\begin{enumerate}
    \item OASYS is an automatic knowledge base construction system that extracts RDF triples from a raw text. It consists of two main components: EL and RE, which fuses deep learning models and structured knowledge information to improve system performance. Moreover, both components can automatically generate labeled training data relying on self-supervised learning and distant supervision. Therefore, OASYS is a domain-agnostic trainable system that requires only a target domain text corpus and a pre-defined KB.
    \\
    \item OASYS is also the first system to build a KB in the Korean language. However, there is no Korean benchmark dataset to evaluate the system. Therefore, we present these new Korean benchmark datasets with a human-annotated test dataset for system evaluation and future research on a variety of Korean NLP tasks. 
\end{enumerate}

%% file: method.tex
\section{System Architecture - OASYS}

\begin{figure*}[h]
    \centering
    \includegraphics[width=\textwidth]{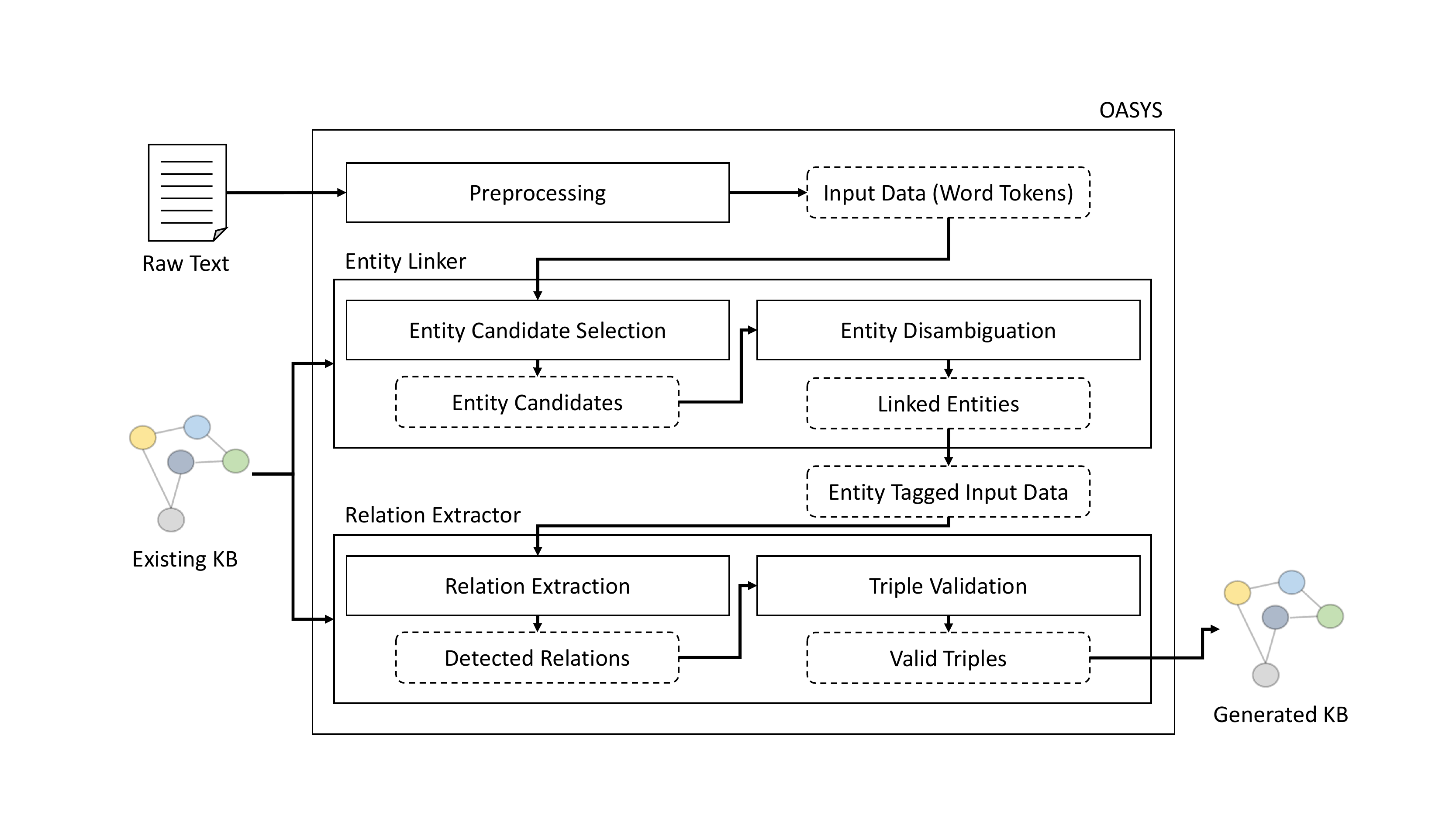}
    \caption{The overview of OASYS.}
    \label{fig:overview}
\end{figure*}

In this section, we describe the entire structure of OASYS architecture. OASYS, the knowledge base construction pipeline system, is a domain-agnostic trainable system, where all processes in the system, containing data generation and model training, are automatically executed. Whenever a demand for the new knowledge base construction system is required, we simply apply the auto-training process and we could construct the new knowledge base construction system. OASYS extracts KB from the text; whenever the new texts like enterprise documents or news articles are published, OASYS links the entities and extracts valid relations from all linked entity pairs in the text.
OASYS consists of two main components: EL and RE. As each component is a delicately designed, data generation and model learning processes can be executed independently. When a raw texts and existing KB are given, each component generates its own appropriate data. EL learns the task of detecting entities in existing KB from raw text through a self-supervised learning technique. RE creates data for training through distant supervision method. To improve the quality of data generated by distant supervision, we first tag the entity information to the corpus using EL, which is trained in advance, and then we generated the data for RE using the entity-tagged text and the existing KB. After both components have been trained, we put them into one single pipeline system, OASYS, which extracted new triples from raw texts. At First, OASYS receives raw text as input, tokenizes it for easy analysis, and adds useful information. This processing contains some tasks such as word tokenizing, POS tagging, dependency parsing and so on. EL takes the parsed input data, finds all entity spans that appear in the text and connects them to unique entity identities in KB. This entity information is appended to the parsed input data and passed to the following component, the RE. RE takes the entity tagged data and extracts valid relation between all entity pairs. The valid triples extracted through OASYS are added to the existing KB to enrich them. The overall OASYS system architecture is depicted in Fig.~\ref{fig:overview}. The following sub-sections describe the detailed architecture of respective components.

\input{preprocessing}
\input{el}
\input{re}

%% file: preprocessing.tex
%\subsection{Preprocessing}
%OASYS performs several preprocessing tasks to use various text information effectively and extracts triples from paragraphs or several sentences as input. For some internal component processes, it is useful to divide the input text into sentences for efficient analysis. We use an open-source sentence splitter KSS~\cite{kss-python} to separate the Korean input text into sentences. After dividing the input text into sentences, all special characters and punctuations are removed from the sentences except for normal words and numbers. Afterward, all sentences are tokenized in word tokens and morphological analysis is performed. Tokenizing, morpheme analysis, and POS tagging are performed using our own NLP tool. RE requires dependency information for each sentence. We use UDify~\cite{kondratyuk201975} parser to analyze the dependency tree of sentences. This preprocessing is identically applied to all input texts used for training and inference of OASYS.

%% file: el.tex
\subsection{Entity Linker}

\indent Before starting this section, we define the recognized phrase of entity name as an entity name span, the unique object in the KB as an entity, and the subset of entities as candidates.\\
\indent This section proposes the first component of OASYS, EL, which predicts an entity name span of phrases. For instance, the given sentence \textit{``Robert Downey Jr. starred in Avengers: Endgame.''}, the model recognizes the phrase \textit{``Robert Downey Jr.''} as an entity name span and selects a subset of entities that is possible to link to \textit{``Robert Downey Jr.''} Afterward, the model computes ranking scores to all elements of sub-set, and the model links the highest score entity, where the result format is like a \textit{Robert Downey Jr.(Person): Q165219.}\\
\indent To resolve above tasks, EL is composed of two sub-components; the \emph{candidates selector} and \emph{entity disambiguation method}.\\
\noindent \textbf{A. Candidates Selector}\\
\indent In the first sub-component, the model recognizes phrases that are possible to entity name spans and selects a subset of entities. The following elements are used to select candidates.
\begin{itemize}
    \item \emph{Entity Recognizer} : OASYS uses conditional random field~\cite{lafferty2001conditional} combined with pre-trained BERT~\cite{devlin2018bert} for entity recognizer. This component recognizes possible phrases to entity name and annotates entity types of phrases - \textit{``Robert Downey Jr.: Person''} and \textit{``Avengers: Endgame: Movie''} etc.
    \item \emph{Word and Entity Joint Embeddings} : In an entity linking system, a unique entity requires to be represented by a unique vector like a word. Therefore, we jointly train words and entities from both the KB and the automatically linked data proposed in Section 3. First, for all 1-hop entity pairs in the KB, we apply node embedding~\cite{grover2016node2vec} to each the neighbor entity pair. 
    \begin{equation}
    \mathcal{L}=\,-{1\over|N(u)|}\,{\sum_{v \in N(u)}}{\textrm{log}\,p(u|v)}
    \end{equation}
    where $u,v$, denote the entity (node) vectors, $N(u)$ denotes all neighborhoods of the node $u$ and $|N(u)|$ denote the number of $N(u)$. Next, from automatically labeled text data, word and pre-trained entity vectors are jointly learned in a single embedding space introduced by~\cite{yamada2017learning}, i.e, entity vectors are represented close to both each 1-hop neighbor entity in the KB and word vectors co-occurring in the same window. Thus, we could train entity representation containing both semantic (from KB) and context (from text) information.
    \begin{equation}
    \mathcal{L}=\,-{1 \over T}\,{\sum_{t=1}^{T}}{\sum_{-c \leq j \leq c, j \neq 0}} {\textrm{log}\,p(v_{t+j}|v_t)}
    \end{equation}
    where, $v$ denotes the word and entity vectors, $T$ denotes the sequence size and $c$ denotes the window size.
    \item \emph{Candidate Generator} : Since the number of entities in KB could be on the scale of millions or billions, OASYS selects candidates for tractable computations: OASYS combines two candidate sets from different selections. One is the selection of all entities having the same entity name spans from the dictionary. The other one is the nearest-neighbors algorithm using pre-trained word and entity joint embeddings, where we compute L2-distance between the current entity name (phrase) and all entities. Afterward, we select k-nearest entities.
\end{itemize}
\noindent \textbf{B. Two-steps Entity Disambiguation Method}\\
\indent The second sub-component computes the ranking score of selected candidates from the entity selector and links the unique entity that achieves the highest score. In a real-world problem, since conventional coarse-grained context-based entity linking models mainly consider entire context information, these models still struggle to use fine-grained relations between entities. For instance, given the sentence \textit{``In 2019, The avengers was released ...''}, entities, the named avengers, exist more than 10 entities in our KB. Therefore, If the more the sentence is long and complex, and the more similar entities exist in the KB, the coarse-grained model, less considering \textit{``released date''} than the entire context, tends to predict noise error results. Thus, we propose both fine-grained that consider pair-wise relations between entities and coarse-grained algorithm; two-steps entity disambiguation method, where the first step is the logical sub-graph based entity linker and the second step is the context-based neural entity linker. These respective models are applied sequentially.
\begin{itemize}
    \item \emph{Sub-graph based neural entity linker} : The first step is the logical brute-force algorithm, in which we construct a sub-graph from all combinations of candidates, and compute the number of respective connections. For the details, we pre-compute the connections between all entities and, after selecting all candidates, we calculate the number of connections between all candidate pairs on the fly. Afterward, if there exists a most connected candidate, we link the most connected candidate and these linked candidates construct a sub-graph.
    \begin{algorithm}
    \caption{Sub-graph based algorithm}
    \label{alg1}
    \KwResult{$e_k(count)$ for all $n$-candidates of entities\;} 
    \textbf{Initialization:} initialize $e_k\textrm{(count)}=0$ ($k \in {1,2,...,n}$)\;
     \For{$i=1$ to $n-1$; $i=i+1$}{
        \For{$j=i+1$ to $n$; $j=j+1$}{
          \eIf{there exists a connection between $e_i$ and $e_j$ in the pre-computed dictionary}{
            $e_i\textrm{(count)} \gets e_i\textrm{(count)}+1$\;
            $e_j\textrm{(count)} \gets e_j\textrm{(count)}+1$\;
           }{
            pass\;
           }
      }
     }
     \eIf{there exists a unique top-1 largest $e_k \textrm{(count)}$}{
        link $e_k$\;
    }{
        execute second-step entity linker\;
    }
    \end{algorithm} If it is possible to link the only most connected entity in the first step, its error rate is negligible and entities are not inferred by the next step. However, since the sub-graph-based algorithm is based on a pre-existing KB, its usability is quite limited. Therefore, we combine the second method.
    \item \emph{Context-based neural entity linker} : The second step computes the ranking score from a given sentence. First, the sentence encoder encodes the sentence.
    \begin{equation}
    v_c=f(c)
    \end{equation}
    where $v_c$ denotes the context vector and $f(x)$ denotes the sentence encoder. We use bi-directional LSTM for sentence encoder $f(x)$. Second, the encoded sentence vector, the word vector of the entity name, and the entity vector of the respective candidates are used to the input of the regression layer, where the regression layer computes the entity-ranking score to link correct entities. 
    \begin{equation}
    s=g([v_c;w_e;v_e])
    \end{equation}
    where $s$ denote final score, $v_c$ denote context vector, $w_e$ denote entity name word vector, $v_e$ denote entity vector, $g(x)$ denote scoring function and $[x;y]$ denote concatenation of x and y. We use 2-layer MLP for scoring function $g(x)$.\\
    This model is only applied for candidates that are not linked from the first stage. The loss function is defined as follows:
    \begin{equation}
    \mathcal{L}=max(0,s'-s+\gamma)
    \end{equation}
    where, $s$ denote a positive score, $s'$ represent a negative score, and $\gamma$ denote the margin of ranking loss. We compute negative score $s'$ from negative entity samples.\\
\end{itemize}

%\begin{figure}[h]
%    \centering
%    \includegraphics[width=80mm]{el.pdf}
%    \caption{The overview of entity linker architecture.}
%    \label{fig:model}
%\end{figure}

%% file: re.tex
\subsection{Relation Extractor}

\indent In this section, we describe the RE. RE is the second component of OASYS, which extracts relational information between entities in a sentence. As the entity information in the sentence is tagged through the entity linker, RE takes the output of the entity linker as input. For example, given the sentence \textit{"Robert Downey Jr. starred in Avengers: Endgame."} and the entity information, RE takes the information and figures out \textit{starred\_in} the relation between \textit{"Robert Downey Jr."} and \textit{"Avengers: Endgame"} entities. A relation such as \textit{starred\_in} is one of the predefined relation sets. For these situations, the relation extraction task can be treated as a multiclass classification problem that classifies the relationship between entity pairs appearing in sentences when the sentences and entity pairs are given as inputs.

\indent To solve the relation extraction problem with the supervised technique, a large amount of labeled data is required for learning. It takes a lot of time and effort to manually annotate these data. To avoid such an exhaustive task, we generate training data automatically through distant supervision~\cite{mintz2009distant} like many other previous models~\cite{zeng2015distant, li2020self, moreira2020distantly, yao2019docred, vashishth2018reside}. Distant supervision automatically generates labeled data from the existing KB and raw text based on the strong assumption that a sentence, in which two entities appear simultaneously, describes the relation of a triple that is connected to the two entities.

\indent However, as the assumption of distant supervision is too strong, generated data can be noisy~\cite{zeng2015distant}. To alleviate this problem, we adopt a multi-instance multi-label learning method~\cite{zeng2015distant, surdeanu2012multi, zhang2019multi}. In the multi-instance learning setting, the assumption of distant supervision is relaxed with at least one of the sentences in which the two entities appear simultaneously describes a target relation. Instead of assigning a single label to one sentence, the label is assigned to a bag of instances (sentences). In addition, tagged labels can be wrong or one bag can be related with multiple labels. Thus, we adopt multi-label learning in which all possible relations for a bag are labeled to the bag. Consequently, RE is trained in a multi-instance multi-label manner. RE consists of two sub-components: relation extraction and triple validation. The following sub-sections describe each sub-component.\\

\noindent \textbf{A. Relation Extraction}

\indent The core logic of RE is based on end-to-end neural network models. Therefore, in this section, we focus more on details of how to construct the model layer-by-layer. We combine various neural network models to extract the semantic relation between two entities appearing in a sentence. The details of neural network layers of relation extraction are described in following paragraph.

\begin{itemize}
    \item \emph{Embedding Layer} :
    RE learns the latent representation of input word tokens simultaneously with learning the relation extraction task. We utilize various lexical features and information to learn the input embeddings for input sentence. There are 4 types of embeddings for each token. They are word, position, entity type, and POS tag embeddings, respectively. First, we used pre-trained Korean GloVe~\cite{pennington2014glove} embeddings for initial word embeddings.  Second, we used position embedding to recognize relative distances from the current word to detected entities. Similar to \cite{zeng2015distant}, we used two relative position embeddings that denote distances between the current token and target entities (subject and object entities). Additionally, we introduced a third position embedding that describes how close the current token is to other entities, except the subject and object, in a sentence. All entities except the target entity pair (subject, object) in the sentence are gathered to composed other entities set $E_o$. If no other entity is detected except for the target entity pair ($|E_o| = 0$), the third position values for all tokens becomes $-1$. However, if other entities exist in the sentence ($|E_o| > 0$), the distance between the i-th word token $w_i$ and all other entities should be calculated. The smallest distance value among them is chosen for the third position value. Lastly, two additional embeddings are used to enrich information about the token. They are POS tag embedding and entity type embedding. These embeddings, except the pre-trained word embeddings, are randomly initialized and learned together.
    
    \item \emph{Feature Layer} : 
    We adopt different models to extract different semantic features from input embeddings.
    As one of the layers extracting vector representations for input sentences, we adopt piecewise convolutional neural networks(PCNN) model ~\cite{zeng2015distant}. PCNN applies convolutional architecture with piecewise max pooling to learn latent feature vectors from given sentences. The convolution output is divided into three segments by subject and object. Piecewise max-pooling is adopted to the segments so that it can capture better structural information between entities than the typical max-pooling method. Formally, PCNN adopts 1D-CNN over the input sequence,
    \begin{equation}
        \textbf{H} = 1D-CNN(\textbf{X}; \textbf{W}^{p}, \textbf{b}^{p})
    \end{equation}
    where, $\textbf{H} \in \mathbb{R}^{d_{h} \times n}$, $\textbf{X} \in \mathbb{R}^{d_{e} \times n}$, $\textbf{W}^{p} \in \mathbb{R}^{d_{h} \times w \times d_{e}}$, and $\textbf{b}^{p} \in \mathbb{R}^{d_{h}}$ are output vectors, input embedding sequence, convolution kernel weight, and bias of 1D-CNN respectively. Then, a piecewise max-pooling performs over the output vectors,
    \begin{align}
    s_{pcnn} =& \nonumber\\
    tanh(&[Pool(\textbf{H}_{1:i});Pool(\textbf{H}_{i+1:j});Pool(\textbf{H}_{j+1:n})]),
    \end{align}
    where $s_{pcnn} \in \mathbb{R}^{3d_{h}}$ is the output vector of the PCNN layer. $H_{1:i}$, $H_{i+1:j}$, and $H_{j+1:n}$ are three segments of the entire convolution output which are divided by $i$ position of subject and $j$ position of object. We apply max-pooling to each sub-sequence. These pooled outputs are concatenated and a non-linear activation function is adopted. As a result, we can get a fixed length result vector $s_{PCNN}$.
    
    For the second vector representation of the input sentence, we utilize the graph convolutional neural networks (GCN)~\cite{kipf2016semi} model described in \cite{zhang2018graph}.
    To use GCN, input sentences should be expressed in graph form. Dependency parsing is applied to express a sentence as a graph. We only use the shortest dependency path (SDP) connecting the two entities in the dependency tree. It enables the model to use the words connecting the subject and object more effectively. The SDP information is represented as an adjacency matrix to be used as the input of GCN. 
    Instead of using a naive adjacency matrix, a self-loop is added to the adjacency matrix and normalized for effective node representation and matching the magnitudes of nodes like \cite{kipf2016semi}. 
    Given an input sequence and its normalized adjacency matrix $\tilde{\textbf{A}}$, $l$-th GCN layer can be expressed as follows.
    \begin{equation}
    h_{i}^{(l)} = \sigma(\sum_{j=1}^{n} \textbf{\~{A}}_{ij}\textbf{W}^{(g_{l})} h_{j}^{l-1} + \textbf{b}^{(g_{l})}),
    \end{equation}
    where $\textbf{W}^{(g_{l})}$, and $\textbf{b}^{(g_{l})}$ are the parameters of $l$-th layer and are updated during training, and $\sigma$ is a non-linear activation function ($ReLU$).
    For more contextualized representation, we apply contextualized GCN (C-GCN) model~\cite{zhang2018graph}. In C-GCN, the word embeddings are first passed through to bi-directional LSTM (Bi-LSTM) to obtain a representation. Then, the output vectors sequence from Bi-LSTM are used as the initial input of GCN layer. Pooling method is applied to extract the final sentence vector from the output vectors obtained from the last layer of GCN. We extract three different representations from different sub-sequences of GCN outputs, entire sequence, subject and object position outputs. It can be formalized,
    \begin{equation}
    h_{sent} = Pool(\textbf{H}_{1:n}),
    \end{equation}
    for whole sentence representation and,
    \begin{equation}
    h_{s} = Pool(\textbf{H}_{s_{start}:s_{end}}),
    \end{equation}
    where $s_{start}$ and $s_{end}$ are the indexes of start and end point of the subject, respectively. The same calculation is applied to object representation $h_{o}$. The final sentence vector is made by concatenating the sentence, subject, and object representation, 
    \begin{equation}
    s_{gcn} = tanh([h_{sent};h_{s};h_{o}]).
    \end{equation}
    
    At last, we apply selective gate method proposed by \cite{li2020self}. Given a sentence bag in which sentences share common entity pairs, first, we extract the hidden vectors ($s_{pcnn}$ and $s_{gcn}$) for each sentence. Relation classification task needs to be performed by aggregating each sentence vector with different importance of information in the bag. 
    The selective gate calculates this importance and returns the weight value between $0$ and $1$.
    We compress the input embedding sequence of each sentence into a single vector by applying the self-attention mechanism to calculate the gating value. First, we calculate the attention weights by the parameterized align function, i.e.,
    \begin{equation}
    \textbf{Q} = \textbf{W}^{(a_{2})} \sigma( \textbf{W}^{(a_{1})} \textbf{X} + \textbf{b}^{(a_{1})} ) + \textbf{b}^{(a_{2})},
    \end{equation}
    \begin{equation}
    \textbf{P} = softmax(\textbf{Q}),
    \end{equation}
    where $\textbf{W}^{(a_{1})}, \textbf{W}^{(a_{2})} \in \mathbb{R}^{d_{h} \times d_{h}}$ and $\textbf{b}^{(a_{1})}, \textbf{b}^{(a_{2})} \in \mathbb{R}^{d_{h}}$ are parameters of self-attention layer. With the attention probabilities, the self-attentive sentence vector is calculated as,
    \begin{equation}
    s_{att} = \sum_{i=1}^{n} \textbf{P}_{i} \odot \textbf{X}_{i},
    \end{equation}
    where, $\odot$ denotes element-wise multiplication.
    Thereafter, a simple feed-forward network is adopted to calculate gating value, i.e.,
    \begin{equation}
    g = \sigma( \textbf{W}^{(g_{2})} \sigma( \textbf{W}^{(g_{1})} s_{att} + \textbf{b}^{(g_{1})} ) + \textbf{b}^{(g_{2})} )
    \end{equation}
    where $\textbf{W}^{(g_{1})} \in \mathbb{R}^{d_{h} \times d_{h}}$ and $  \textbf{W}^{(g_{2})} \in \mathbb{R}^{d_{h} \times 6d_{h}}$.
    Using calculated hidden vectors ($s_{pcnn}$ and $s_{gcn}$) and gating values ($g$) as weight, we can perform weighted sum for the final representation of the input $c$size-bag i.e.,
    \begin{equation}
    s = [s_{pcnn};s_{gcn}]
    \end{equation}
    \begin{equation}
    v = \sum_{i=1}^{c} \textbf{G}_{i} \odot \textbf{S}_{i}
    \end{equation}
    
    \item \emph{Prediction Layer} : 
    RE is also intended to enable multi-labeled inference for a given input bag. After projecting the previously calculated bag vector into the relation label space and converting it into a probability distribution by applying softmax function, we can classify one bag into a relation. Instead of the aforementioned method, the score for each relation label is calculated by a simple MLP layer to the bag vector. Formally, the relation scores are calculated as,
    \begin{equation}
    r = \sigma( \textbf{W}^{(o_{2})} \sigma( \textbf{W}^{(o_{1})} s + \textbf{b}^{(o_{1})} ) + \textbf{b}^{(o_{2})} )
    \end{equation}
    where $\textbf{W}^{(o_{1})} \in \mathbb{R}^{6d_{h} \times 3d_{h}}$ and $  \textbf{W}^{(o_{2})} \in \mathbb{R}^{3d_{h} \times r}$.
    For all relation scores, relation scores $r_{j}$ that exceed the threshold become the predicted relations. If all the scores of relations do not exceed the threshold, there is no relation for the corresponding bag.
    We apply the sliding margin loss~\cite{zhang2019multi} to learn the threshold through training instead of using the fixed value threshold. Loss function for $i$-th relation $r_{i}$ is as follows,
    \begin{align}
    L_{i} &= Y_{i} max( 0, (B + \gamma) - ||r_{i}|| )^{2} + \nonumber \\
    &\quad \lambda(1 - Y_{i}) max( 0, ||r_{i}|| - (B - \gamma))^{2},
    \end{align}
    where $Y_{i}=1$ if the bag represents the target relation $r_{i}$ and $Y_{i}=0$ if the bag is not related to the target relation.
    In the loss, $B$ is the learnable threshold variable. $\gamma$ is a margin value that widen the difference between positive and negative scores. $\lambda$ is the down weighting value applied to the negative loss.
    
\end{itemize}

\noindent \textbf{B. Triple Validation} \\
\indent The triples extracted through relation extraction can be imperfect. Relation extraction is based on some information about the entity and word embedding of the given input sentence. Thus, it is not easy to distinguish whether an inferred triple is true. Nevertheless, it is possible to prevent cases that result in non-sense triples. A non-sense triple refers to the kind of triple in which the subject or object type cannot be connected to a relation of the triple. For example, a triple connected by a relation \textit{starred} commonly has \textit{Work} type entity as a subject and \textit{Agent} type entity as an object. It does not make sense for an \textit{Agent} type to appear in the subject or \textit{Work} type to appear in the object of triple which is linked by \textit{starred} relation. To increase the relation extraction precision, we introduce a simple triple validation that can filter out non-sense triples. For triples in KB, the types of subject and object that can be located can be limited according to the relation type of the triples. Therefore, we can know in advance whether the types of subjects and objects are suitable for the relation by the existing KB. We collect all existing triples connected to each relation to compose the subject-type set and object-type set. We call this  fact-type templates. When triples are extracted through RE, the model checks the types of subjects and objects in it. If the types of the inferred triple do not exist in the fact type templates, the triple is abandoned from the inferred result. With this simple triple validation process, we can avoid extracting completely incorrect triples.

%\begin{figure}[h]
%    \centering
%    \includegraphics[width=80mm]{re_figure.pdf}
%    \caption{Overview of Relation Extractor}
%    \label{fig:re_model}
%\end{figure}

%% file: dataset.tex
\section{Data Construction}
\input{autodata}
\input{humandata}

%% file: autodata.tex
\subsection{Auto Generated Dataset}
\begin{figure}[h]
    \caption{The number of extracted sentences per iteration of entity linking}
    \label{fig:data_extraction}
    \centering
    \includegraphics[width=80mm]{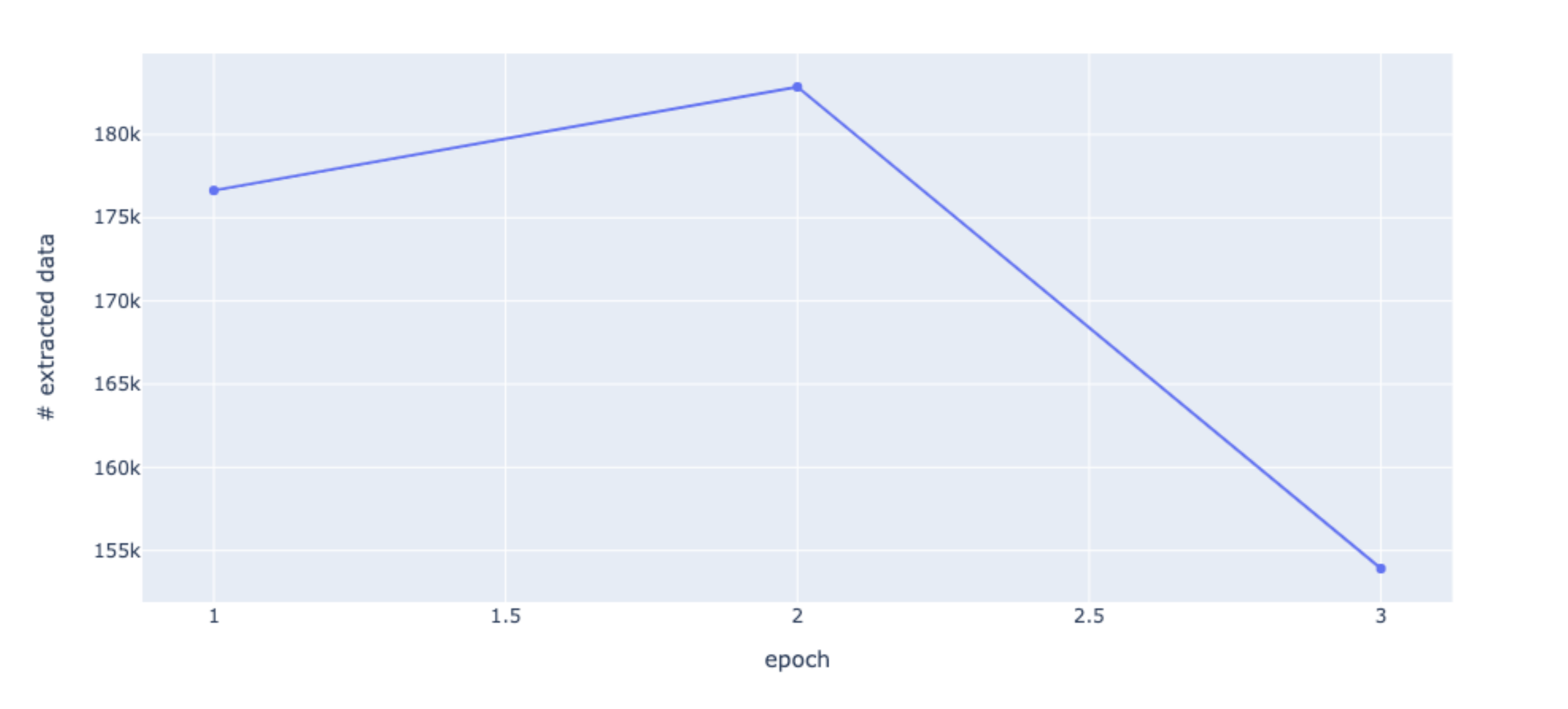}
\end{figure}

\indent OASYS automatically extracts labeled data using self-supervised learning and distant supervision. Algorithm~\ref{alg2} summarizes the process of auto-generation of training data, which extracts triples from raw text.\\
\indent First, we discuss the initialization step that trains an initial entity recognizer from noisily tagged data by longest n-gram matching, where we compare all n-grams to our entity gazetteer and tag the longest name-matched n-grams. Next, to link unique entities from sentences, we apply self-supervised learning, where the current entity recognizer annotates entities, and the sub-graph-based entity linker (details in Algorithm~\ref{alg1}) links the correct unique entities. Thereafter, we filter un-linked data and re-train student entity recognizer using that data (labeled from teacher model). As can be seen from Table~\ref{table3}, since the sub-graph-based linker 
obtains significantly outstanding precision (error rate 0.6\%) in human-labeled test data, we can extract the clean annotated dataset. Finally, We repeat this process until the amount of extracted data decreases or iterations\footnote{In data construction, we use both terms iteration and epoch for entire 1-step.} larger than 3. We realize that if the amount of extracted data decreases, the current entity recognizer 
falls into over-fitting, and the previous model achieves the best performance. Moreover, in most domain use-cases, the amount of extracted data decreases before 4 iterations. Fig.~\ref{fig:data_extraction} also illustrates that the best model is converged at 2 iterations. After these iterations for entity linking, we generate data for RE by distant supervision with existing KB and entity tagged sentences. Each sample of auto-generated data consists of an entity tagged sentence and its associated triple. We randomly split auto-generated data into training, validation, and test sets. \\
\indent In this study, we present auto-generated data that is extracted from the entire Korean Wikipedia corpus and sub-set of the Korean DBpedia~\cite{lehmann2015dbpedia} triple set. A sub-set of the Korean DBpedia is extracted with several criteria which are used in human-annotated test data (details in Section 3.2). The size of this automatically generated dataset is shown in Table~\ref{table1}.
\begin{table}[h]
    \centering
    \caption{Size of datasets}
    \label{table1}
    \begin{tabular}{c|c|c|c}
         & training set & validation set & test set\\
         \hline
         \textbf{Auto-generated} & 366,872 & 45,961 & 45,962\\
         \textbf{Human-labeled} & $\cdot$ & $\cdot$ & 6,058\\
    \end{tabular}
\end{table}

\begin{algorithm}[h]
\caption{Auto-generated training data process}
\label{alg2}
   \SetAlgoLined
    \KwData{
    Noisily span tagged text data by longest n-gram entity name matching}
    \KwResult{
    An automatically generated data}
    \textbf{Initialization:} Train an initial entity recognizer from noisily tagged data; \\
    \While{the number of extracted data is increasing and the total number of steps is less than 4}{
        1. Recognize entities by a previously trained model.\\
        2. Link unique entities using by Sub-graph based linker and Filter un-linked data.\\
        3. Train a new student entity recognizer by filtered data from above 2.\\
    }
    4. Apply distant supervision between entity tagged text and DBpedia KB.
\end{algorithm}

%% file: humandata.tex
\subsection{Human-Labeled Test Dataset}

OASYS was trained using auto-generated data and evaluated on auto-generated test data. Although results of the auto-generated evaluation dataset can be used to estimate the performance of trained systems, we also consider a human-annotated test dataset for accurate and general evaluation of the systems. Therefore, we introduce a new benchmark dataset to evaluate the task of triple generation. This dataset is generated through human labeling and based on Korean DBpedia and Korean Wiki Article data.

We wanted our test dataset to be able to make general evaluations. For this reason, several criteria were established and we used only data that met these criteria. We filtered the Korean DBpedia triples used in the data generation using the following conditions. First, we used triples whose subject and object URLs start with \textit{'http://ko.dbpedia.org/'} to consider only triples that bridge entity to entity. For relations, we used only relations starting with \textit{'http://ko.dbpedia.org/'}. Thereafter, we grouped entities by the entity type and collected relations directly connected to the top-50 entities with the most connectivity within each entity group to form a relation set. Finally, only relations with the occurrence of 2 or more were selected, and we acquired 156 relations in the created dataset. There were about 290,000 triples that satisfy these entity and relation conditions. These triples were also used to auto-generate training data. We also choose Wiki articles with rich connectivity. Among the articles linked to DBpedia entity, the top 1,147 articles with most connections were selected. The connectivity was calculated by counting edges whose relation belong to previous 156 relation set. Brief statistics of used DBpedia triples are summarized in Table \ref{table2}.
The test dataset was compiled by educated human annotators labeling the sentences of wiki articles one by one. Annotators looked at each sentence and checked for the subject and object entities and if there was a triple that describe the two entities, they tagged the information. Through this process 6,058 of (sentence, triple) pair data were generated (Table \ref{table1}).

\begin{table}[h]
    \centering
    \caption{DBpedia triples used in this paper}
    \label{table2}
    \begin{tabular}{c|c|c}
         & Entire DBpedia Dump & Filtered DBpedia\\
         \hline
         \textbf{Entities} & 9,881,982 & 108,280\\
         \textbf{Relations} & 16,595 & 156\\
         \textbf{Triples} & 27,956,860 & 291,215\\
    \end{tabular}
\end{table}

%% file: experiment.tex
\section{Result}
OASYS is evaluated on the test data from both auto-generated data and human-labeled data since the validating of the general applicability of OASYS is necessary. In other words, as the auto-generated test data inevitably contains incorrectly labeled data, we require to test on clean test data, although we train only auto-generated training data. Therefore, we construct human-labeled test data which is similar to real-world data and focus on the model performances to this dataset. We test sub-components respectively and detailed experimental settings are described in each sub-section.
\input{el_result}
\input{re_result}

%% file: el_result.tex
\subsection{Results of Entity Linker}
This section evaluates the performance of entity linking. For detailed analysis, we test sub-graph based linker and context-based linker respectively.

\begin{table}[h]
    \centering
    \caption{The top-1 precision and coverage of possible linking of the sub-graph based linker}
    \begin{tabular}{c|c}
         Error rate (Precision@1) & Coverage\\
         \hline
         0.6\% (99.4) & 60.2
    \end{tabular}
    \label{table3}
\end{table}
\indent For evaluating sub-graph based linker, we evaluate this algorithm to a human-labeled test set. Table~\ref{table3} shows the Top-1 Precision and Coverage of possible linking of this algorithm. Since this method only links possible entities, we only evaluate Error rate (Top-1 Precision). This result demonstrates that the fine-grained sub-graph based algorithm obtains a negligible error rate, thus, we could extract the more clean triple sets. Moreover, the coverage illustrates that this algorithm links about 60\% of entities in the test set.\\
\begin{table}[h]
    \centering
    \caption{The evaluation results of the context-based linker to both auto-generated and human-labeled test sets}
    \label{table4}
    \begin{tabular}{c|c|c}
         & Accuracy@1 & Mean rank\\
         \hline
         \textbf{Auto-generated} & 86.5 & 1.32\\
         \textbf{Human-labeled} & 89.1 & 1.19
    \end{tabular}
\end{table}\\
\indent Since Context-based neural entity linker is trained by only an automatically generated training set, and since we search the hyperparameters on the automatically generated validation set, we evaluate both automatically generated test set and human-labeled test set for measuring the generalization and robustness of unseen distributions. To evaluate Context-based neural entity linker, since the entity linking is a ranking problem, we evaluate the Top-1 Accuracy (the entity acquiring the rank-1 score is true) and Mean Rank (average rank of true rank-1 entities predicted by this method). Table~\ref{table4} shows the experimental results on the automatically generated test set and human-labeled test sets respectively. Notably, this method obtains higher performance in the human-labeled test set than the automatically generated test set. This result demonstrates that entity linker is robust to real-world problem.

%% file: re_result.tex
\subsection{Results of Relation Extractor}
In this section, we evaluate the performance of RE. 
The examined RE is trained on only auto-generated data from distant supervision. 
We evaluate RE on both auto-generated and human-labeled test data. The auto-generated test data could be imperfect and inherently contains some noise. For objective evaluation, we evaluate our system on the both datasets. The number of data samples in auto-generated and human-labeled test data set are 45,962 and 6,058, respectively. The result is illustrated in the Table~\ref{re_result_table}. The datasets widely used for relation extraction~\cite{riedel2010modeling, zeng2018extracting, nayak2020effective, takanobu2019hierarchical} have been provided. However, they do not contain a human-labeled test set or contain a limited number of samples and small sized relations comparing to our dataset. Therefore, the performance results of the model shown on provided dataset are meaningful. From this result, it can be concluded that the generated KB from OASYS trained on only uto-generated data is useful.

\begin{table}[h]
    \centering
    \caption{Evaluation results of RE on both auto-generated and human-labeled test sets}
    \label{re_result_table}
    \begin{tabular}{c|c|c|c|c}
         Dataset & Accuracy & Precision & Recall & F1-Score\\
         \hline
         \textbf{Auto-generated} & 80.4 & 77.2 & 70.4 & 73.7 \\
         \textbf{Human-labeled} & 57.8 & 60.4 & 55.5 & 57.8 \\
    \end{tabular}
\end{table}

%% file: conclusion.tex
\section{Conclusion}
We propose a domain-agnostic trainable knowledge base auto construction system (OASYS), which trains with self-supervised learning and distant supervision. Moreover, Although OASYS trains without human-labeled data, experimental results illustrate that its performance on human-labeled test data is useful. In general, as knowledge base auto-construction systems struggle due to noise prediction results, we combine several sub-components to reduce errors in prediction results. We are the first to propose the Korean-language knowledge base auto-construction system and provide a benchmark dataset with large quantity and various relations types for reproducing models and supporting many researchers who are interested in the Korean-language knowledge extraction task. In the future, we will develop our system not only extract relations from pre-defined KB but also extract relations that do not exist in KB. We also plan to integrate the currently separated learning processes (EL and RE), where the entire system can be learned in an end-to-end manner. We are continuously trying to develop and elaborate our system while applying it to many possible use cases.